\PassOptionsToPackage{table}{xcolor}
\documentclass{article}
\usepackage{iclr2027_conference,times}
\usepackage[T1]{fontenc}
\usepackage{amsmath,amsfonts,bm}

\def\eqref#1{equation~\ref{#1}}
\def\1{\bm{1}}

\DeclareMathAlphabet{\mathsfit}{\encodingdefault}{\sfdefault}{m}{sl}
\SetMathAlphabet{\mathsfit}{bold}{\encodingdefault}{\sfdefault}{bx}{n}

\usepackage[hyphens]{url}
\usepackage{graphicx}
\graphicspath{{figures/}{./}}
\usepackage{booktabs}
\usepackage{amsfonts}
\usepackage{amsmath}
\usepackage{amssymb}
\usepackage{enumitem}
\usepackage{algorithm}
\usepackage{algpseudocode}
\usepackage{xcolor}
\usepackage{colortbl}
\usepackage{placeins}
\usepackage{hyperref}
\hypersetup{colorlinks=true,citecolor=black,linkcolor=blue,urlcolor=blue}

\setlist[itemize]{leftmargin=*}
\setlist[enumerate]{leftmargin=*}

\definecolor{tabheader}{RGB}{214,227,240}
\definecolor{tabbest}{RGB}{198,230,200}
\definecolor{tabalt}{RGB}{244,246,248}

\title{COTCAgent: Preventive Consultation via Probabilistic Chain-of-Thought Completion}

\author{
\textbf{Zihan Deng}$^{1,\dagger}$\thanks{Equal contribution.}\thanks{Corresponding author. \protect\href{mailto:zihandeng@connect.hku.hk}{\textcolor{black}{\protect\raisebox{-0.12em}{\protect\includegraphics[height=0.95em]{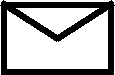}}}}},
\textbf{Chuanzhi Xu}$^{2,\dagger}$\footnotemark[1],
\textbf{Xiaozhen Zhong}$^{3}$,
\textbf{Quankeng Huang}$^{1}$\\[0.45em]
$^{1}$The University of Hong Kong\\
$^{2}$The University of Sydney\\
$^{3}$University of Electronic Science and Technology of China\\[0.5em]
\href{https://github.com/FrankDengAI/COTCAgent}{\textcolor{black}{\raisebox{-0.28em}{\includegraphics[height=1.45em]{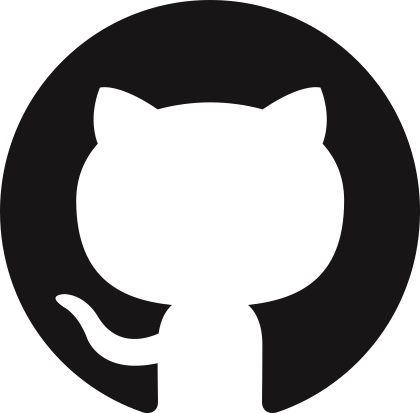}}}}
}

\iclrfinalcopy
\makeatletter
\def\@maketitle{%
 \vbox{\hsize\textwidth
  \centering
  {\LARGE\bfseries \@title\par}
  \vskip 0.75em
  {\large \@author\par}
  \vskip 0.28in
 }%
}
\makeatother

\begin{document}

\maketitle
\fancyhead{}
\renewcommand{\headrulewidth}{0pt}
\pagestyle{plain}

\begin{abstract}

  Sequential diagnosis requires ranking diseases from a sparse intake under a public budget of follow-up questions, which is diagnostic triage with missing findings rather than screening of people who have no symptoms.
  Because laboratory series arrive irregularly and histories remain incomplete, the ranking must be updated as new facts arrive, yet most pipelines are limited by unnamed numeric trends, unverifiable free-form chain of thought, and a negative bias from treating unasked findings as absent.
  To solve these, we propose COTCAgent, a sparse-intake consultation agent built on Chain-of-Thought Completion (COTC), which couples three modules on one shared ternary evidence log.
  The Temporal-Statistics Adapter (TSA) turns time series into typed predicates with named statistics, while the COTC module scores diseases with a learnable knowledge-base energy in which unknown findings add zero energy, after which bounded probabilistic completion selects unknown findings under the question budget and writes each answer back as present, absent, or unknown.
  Edge strengths and an absence scale are trained with listwise ranking under a sparse curriculum, so that the same energy chooses the next finding by a surrogate entropy reduction.
  Experiments on DDXPlus and MIMIC-IV show that these modules improve ranking under a matched question budget, with information-gain completion above asking nothing or at random and above protocol-matched askers and a local LLM, thereby providing a foundation for future calibrated scoring and prospective preventive consultation.
\end{abstract}

\section{Introduction}
  \label{sec:intro}
  
  Sequential diagnosis is hard because labs arrive at irregular times, histories are incomplete, and the ranked list of diseases must change as new facts arrive~\cite{zhou2025large,wornow2024context,nori2025sequential}.
  In sequential diagnosis, preventive consultation ranks diseases from a sparse intake: the agent may ask a public number of follow-up questions and then returns a ranked list. The setting is triage with missing findings, not screening of people with no symptoms. DDXPlus cases and MIMIC-IV admissions are already symptomatic or already in hospital.
  The practical questions are which unknown finding to ask next so that the ranked list improves, and how each update can be checked against a named computation rather than against free-form text.
  
  Language models are widely used for medical dialogue, monitoring, and tool-using workflows~\cite{singhal2023towards,wang2024agentbench,liu2024toolllm}. Three limitations recur.
  First, numeric trends are often written in words with no named estimator, so slopes and change points are hard to check.
  Second, free-form chain of thought can draft questions, but the resulting trace is hard to check against a structured knowledge base.
  Third, many interactive pipelines turn incomplete charts into full symptom vectors and mark every unasked finding as absent. That pattern appears in older tabular classifiers and RL symptom checkers, and also in LLM stacks that fill unmentioned findings as negative. The result is a strong negative bias, a penalty on incomplete intakes, and little value to asking.
  
  Medical LLMs were often tested on templated intakes or multiple-choice items that do not look like chart review~\cite{singhal2023towards,mcduff2023capabilities}.
  Longer context windows change how much of a record fits in a prompt~\cite{liu2025comprehensive,wornow2024context}. They do not by themselves give faithful numeric reasoning or questioning under a budget.
  Tool-using and multi-agent stacks move arithmetic and retrieval out of the decoder~\cite{wang2024agentbench,luo2025agentmath,liu2024toolllm}. Agents that read long records can still invent trend claims and rationales that are hard to map back to knowledge-base rows~\mbox{\cite{feli2025llmagent}}.
  We therefore focus on preventive consultation under a matched question budget: complete a structured evidence log, re-rank diseases with a score that can be replayed, and report both Hit@1 and question cost so that policies can be compared.
  
  We do not optimize open-domain dialogue or free-form clinical notes. We ask whether a small knowledge-grounded scorer can turn sparse intakes into stronger ranked lists when every follow-up question is counted and every answer maps into a finite evidence vocabulary.
  
\begin{figure}[t]
    \centering
    \includegraphics[width=1.03\linewidth]{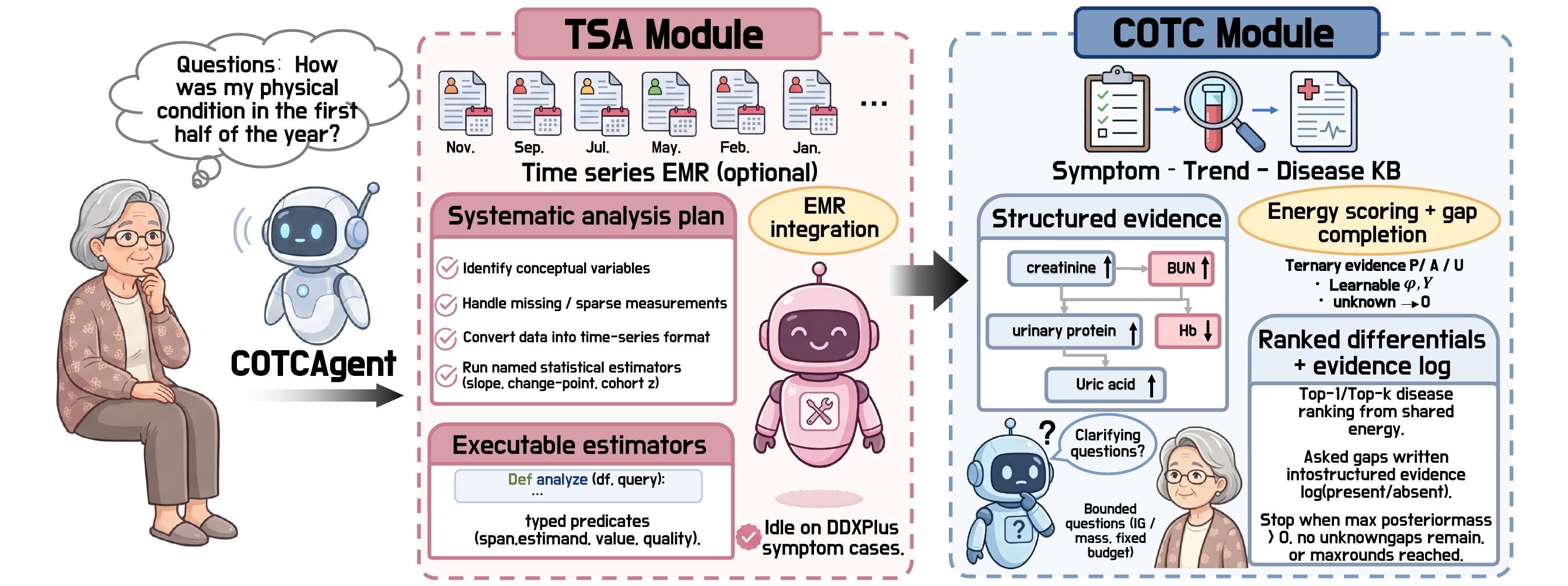}
    \caption{COTCAgent overview. TSA emits typed trend predicates from time series when those series exist.
    The COTC module scores diseases and drives bounded completion over ternary evidence.}
    \label{fig:architecture}
  \end{figure}

  We propose COTCAgent in Section~\ref{sec:method} and Algorithm~\ref{alg:cotc_loop} as three layers, shown in Figure~\ref{fig:architecture}.
  The Temporal-Statistics Adapter turns time series into typed trend predicates with executable estimators when such series exist.
  The Chain-of-Thought Completion module, abbreviated COTC, scores diseases with inverse-disease-frequency weights and a learnable edge matrix $\phi$ over a symptom, trend, and disease knowledge base. This module is the only part trained from labeled differentials.
  Bounded probabilistic CoT completion then asks only about unknown findings, parses short answers into ternary evidence, and stops when a mass threshold is met, when no unknown finding remains, or when the round limit is reached.
  Chain of thought here is the growing evidence log and the energy trail, not token-level reasoning inside a language model. The loop is iterative completion over a finite vocabulary.
  We restrict the agent to templated questions so that each step stays tied to a knowledge-base row and a dialog turn. Our contributions can be summarized as follows:
  \begin{itemize}
  \item We propose COTCAgent, a sparse-intake consultation agent that couples executable TSA predicates, a learnable COTC scoring module, and bounded probabilistic CoT completion on one shared ternary evidence log. Unknown findings add zero energy, and present or absent findings update a knowledge-base energy that can be replayed.
  \item We introduce listwise training of edge strengths $\phi$ and an absence scale $\gamma$ under a sparse curriculum. The same trained energy is reused for ranking and for follow-up selection, so each update stays tied to a knowledge-base row rather than to free-form text.
  \item We evaluate matched-budget consultation on DDXPlus and MIMIC-IV. Information-gain completion reaches 88.1\% Hit@1 with 3.61 questions on DDXPlus and lifts phenotype Hit@1 from 72.7\% to 79.4\% on MIMIC-IV. Protocol-matched copies of recent acquisition methods and a local LLM that both asks and ranks stay below learned COTC under a shared answer oracle.
  \end{itemize}
  
  \section{Related Work}
  
  Medical LLMs, Tool Agents, and Sequential Diagnosis.
  Medical LLMs have moved from static QA and capability tests~\cite{singhal2023towards,mcduff2023capabilities,esteva2019guide} toward long-context EHR modeling~\cite{wornow2024context,liu2025comprehensive}.
  Agent benchmarks move planning, retrieval, and code outside the decoder~\cite{wang2024agentbench,liu2024toolllm,luo2025agentmath}. Multi-agent stacks widen collaborative diagnosis~\cite{qian2023communicative}. Sequential protocols stress coordinated questioning~\cite{nori2025sequential,zhou2025large}.
  Many systems still mix drafting, evidence updates, and ranking in free-form generation. Cost and checkability are then hard to see under sparse intakes.
  COTCAgent keeps language at template questions and short-answer parsing, so the COTC module stays an explicit energy with a public question cost. We compare against a local LLM that both asks and ranks in Table~\ref{tab:ext_baselines}.
  
  Knowledge-Grounded and Neuro-Symbolic Medical Reasoning.
  A second line grounds predictors in graphs or rules~\cite{xu2025survey,wang2023biobridge,delong2023neurosymbolic,garcia2025nssc}.
  Symptom checkers often use fixed likelihood tables. Those tables are easy to inspect but hard to adapt when edges should be learned from labels.
  On DDXPlus, BASD, AARLC, and Diaformer are standard interactive baselines~\cite{fansi2022ddxplus,yuan2021aarlc,luo2020asd,chen2022diaformer}.
  COTCAgent learns continuous edge strengths on a discrete incidence matrix, keeps unknown slots explicit, and does not invent knowledge-base relations when no unknown finding remains.
  
  Probabilistic Scoring, Missingness, and Active Acquisition.
  Graphical models give a clear meaning when priors exist~\cite{koller2009probabilistic}. Many ranked lists still treat unseen findings as absent.
  Ternary present, absent, and unknown logs are already standard in interactive symptom checkers~\cite{fansi2022ddxplus,yuan2021aarlc}. We keep that encoding so unknown findings add zero energy under listwise training.
  Expected information gain is classical experimental design~\cite{guan2021bed}. Because our scorer is discriminative rather than generative, we use a one-step information-gain surrogate on the trained ranking scores $\tilde{P}$, not calibrated mutual information from a symptom likelihood and not token-level LLM uncertainty.
  Each update writes a ternary token onto a knowledge-base incidence row and adds to a public question cost.
  We compare information-gain completion to mass, frequency, fixed-KB BED, and learned neural askers under the same answer oracle.
  
  \section{Method}
  \label{sec:method}
  
  \subsection{Overview of COTCAgent}
  
\begin{figure}[t]
    \centering
    \includegraphics[width=\linewidth]{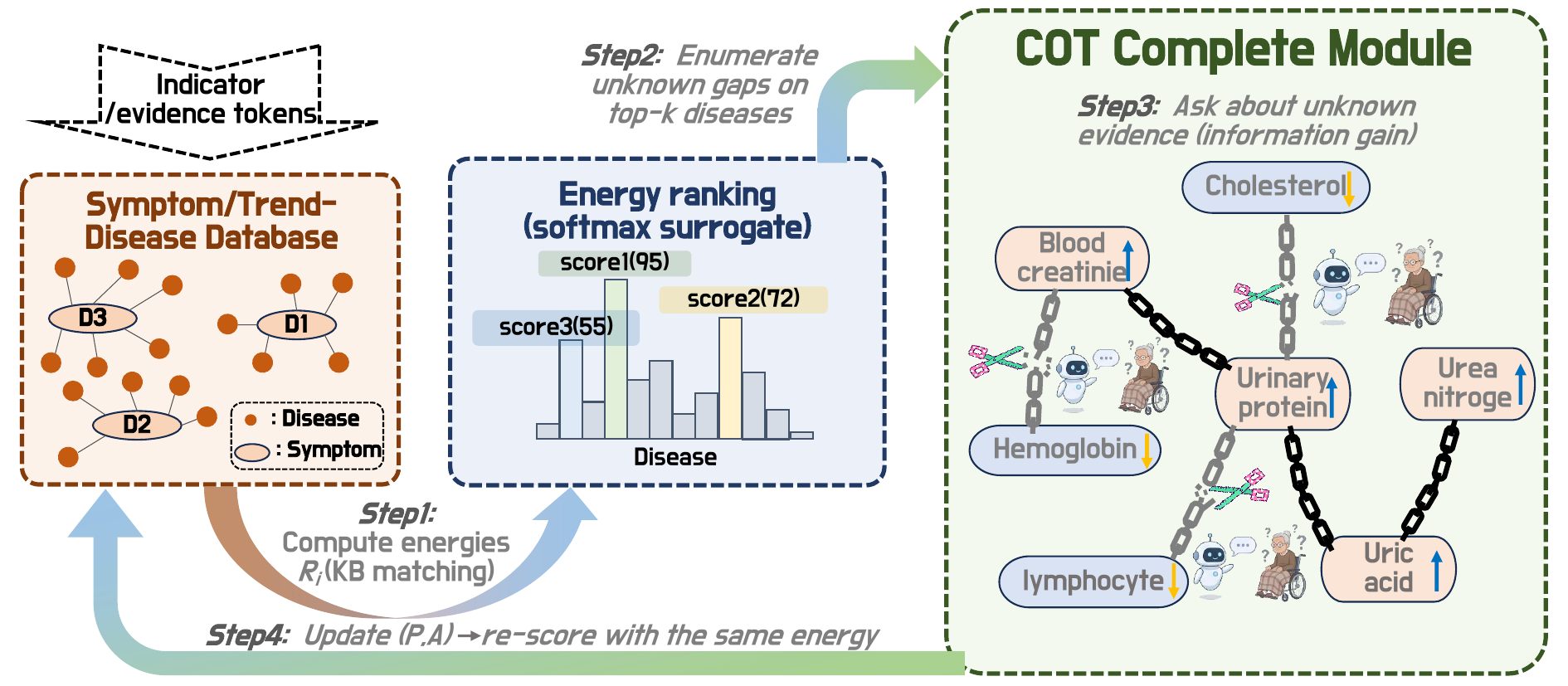}
    \caption{COTC scoring and completion.
    Structured cues feed the knowledge-base matcher. Ranked diseases, explicit unknown findings, and follow-up turns share the same energy in Algorithm~\ref{alg:cotc_loop}.}
    \label{fig:pipeline}
  \end{figure}

  COTCAgent ranks diseases with optional follow-up questions, as shown in Figure~\ref{fig:architecture}.
  TSA produces predicates from labs and vitals when time series exist.
  The COTC module matches symptoms and predicates against a knowledge base and returns a ranked list with explicit unknown findings, as in Figure~\ref{fig:pipeline}.
  Bounded completion fills those unknown findings until a stopping rule fires.
  Natural language is used only to render templated questions and to parse short answers into discrete evidence.
  
  We keep three surfaces separate so that failures stay easy to see.
  Executable statistics consume raw series under named likelihoods, so slope and change-point claims are regressions rather than free-text guesses.
  The COTC module consumes only predicates and evidence tokens that belong to a finite schema.
  Natural language only paraphrases those pre-specified unknown findings and parses short answers back into the same tokens.
  When TSA fails, the COTC module receives an explicit uncertainty flag instead of an invented numeric claim.
  When no unknown finding remains or the mass threshold is met, Algorithm~\ref{alg:cotc_loop} returns the current ranking and does not invent knowledge-base edges.
  
  \subsection{Temporal-Statistics Adapter}
  \label{sec:tsa}
  
  TSA maps a clinical analysis request $Q$ to typed predicates through a deterministic adapter
  \begin{equation}
  \label{eq:tsa_map}
  f_{\mathrm{TSA}}: Q \rightarrow \mathcal{I} \rightarrow \Lambda \rightarrow C,
  \end{equation}
  where $\mathcal{I}$ is a rule-parsed intent (which series and which target statistic), $\Lambda$ is a named estimator family, and $C$ is the executable routine that implements that family.
  The request $Q$ is not free-form user chat. It is a structured tuple of series name, timestamps, values, and a target key such as slope or change-point.
  The planner inside TSA is a lookup from that key to a routine. It is not a language-model planner, and it does not generate code at evaluation time.
  Trend queries fit an ordinary least-squares slope on the current patient's series,
  \begin{equation}
  \label{eq:mixed}
  y_{ij}=\beta_0+\beta_1 t_{ij}+u_i+\epsilon_{ij},\qquad u_i\sim\mathcal{N}(0,\sigma_u^2),\ \epsilon_{ij}\sim\mathcal{N}(0,\sigma_\epsilon^2),
  \end{equation}
  where $i$ indexes the patient, $j$ the observation, and $t_{ij}$ time. Equation~\ref{eq:mixed} records the cohort-level random-intercept extension for settings in which repeated patients are modeled jointly; the online single-patient TSA uses the OLS special case $u_i=0$ and does not claim to estimate $\sigma_u^2$ from one trajectory.
  When a training cohort is available, as on PhysioNet, TSA also attaches a cohort $z$-score that uses the training-split mean and variance of the last observation.
  Abrupt-change queries scan a split index $\tau$ by reduction in residual sum of squares,
  \begin{equation}
  \label{eq:changepoint}
  \mathrm{score}(\tau)=\mathrm{SSE}(\mathbf{y})-\big(\mathrm{SSE}(\mathbf{y}_{1:\tau})+\mathrm{SSE}(\mathbf{y}_{\tau+1:T})\big),\qquad \tau\in\{2,\ldots,T-2\},
  \end{equation}
  which matches a Gaussian two-piece mean-shift model with a uniform prior on $\tau$.
  Failed fits fall back to simpler diagnostics and attach SPARSE or UNSTABLE flags. Instability is residual RMSE relative to the series mean, so the COTC module can down-weight the predicate.
  Each TSA artifact is a typed tuple of span, target statistic, value, and quality drawn from a finite schema that covers slope, change-point score, and cohort $z$. Residual magnitude is a quality flag, not a separate evidence token.
  Numeric claims stay tied to named routines. The planner only selects which target statistic to run, and the returned object is always the output of that documented procedure.
  
  On DDXPlus, the temporal estimators are unnecessary because cases already provide discrete symptom evidence, but TSA is not bypassed: its schema and validation path map released fields into the same typed present/absent/unknown predicates consumed by COTC. Thus DDXPlus exercises typed evidence normalization while not exercising slope or change-point estimation.
  On MIMIC-IV, TSA consumes irregular chemistry series and emits the discrete abnormal and trend predicates used by COTC, including the deterministic lab bins and available longitudinal summaries. The reported MIMIC ranking and completion results therefore use TSA-derived evidence tokens; they do not fit a cohort random-intercept model at test time.
  PhysioNet-style checks of which statistics are emitted appear in Appendix~\ref{app:physio}.
  
  \subsection{COTC Module: Ternary Energies and Learning}
  \label{sec:cotc_energies}
  
  The COTC module is the trainable core of COTCAgent.
  We do not train a language-model reader. We do not train the question policy.
  We train only the knowledge-base edge strengths $\phi$ and the absence scale $\gamma$ that define the ternary energy below, under listwise ranking and a sparse curriculum.
  
  Knowledge Base.
  We use a symptom, trend, and disease incidence matrix $M\in\{0,1\}^{|D|\times|E|}$, distinct from the TSA adapter $f_{\mathrm{TSA}}$ in Eq.~\ref{eq:tsa_map}.
  On DDXPlus we instantiate $D$ and $E$ from the public release with 49 pathologies and 223 evidences.
  On MIMIC-IV we build a co-occurrence knowledge base between phenotype groups and lab-derived evidence tokens on the training split only.
  Edges that never appear in training remain at their initialization and do not receive gradients.
  
  Inverse Disease Frequency as a Fixed Sufficient Statistic.
  Common findings appear in many diseases and are weakly discriminative. Rare findings shrink the active set faster.
  For evidence $e_j$ we define an IDF weight that depends only on knowledge-base incidence,
  \begin{equation}
  \label{eq:idf_weight}
  w_j = \log\frac{|D|+1}{|\{d_i: M_{ij}=1\}|+1}.
  \end{equation}
  The outer $\log w_j$ in Eq.~\ref{eq:diag_score} is extra flattening, not a second copy of IDF. The weight $w_j$ already compresses incidence, and $\log w_j$ further flattens common findings.
  We clamp $w_j\ge 1$ before that outer log so that any finding present in more than about $1/e$ of diseases contributes $0$ from the IDF term ($\log 1=0$) and is carried only by $\log\phi_{ij}$.
  This clamp is strong on dense co-occurrence bases such as MIMIC phenotype chapters. Without the clamp, $\log w_j<0$ would invert ranking toward empty rare diseases.
  When population prevalences are unavailable, $w_j$ acts as a surrogate log-odds weight that can be checked against a knowledge-base row and does not need a generative symptom model.
  Holding $w_j$ fixed while learning $\phi$ separates two roles. IDF encodes how rare a finding is in the ontology. The matrix $\phi$ encodes how strongly that finding should support a particular disease after seeing labeled differentials.
  
  Ternary Evidence and Additive Gibbs Energy.
  Let $P,A\in\{0,1\}^{|E|}$ be masks for present and absent evidences, and define unknown slots by $U=\mathbf{1}-P-A$, with present preferred if both flags are set.
  Only known facts update the energy:
  \begin{align}
  \label{eq:diag_score}
  R_i &=
  \sum_{j: M_{ij}=1} P_j \big(\log w_j + \log \phi_{ij}\big)
  - \sum_{j: M_{ij}=1} A_j\,\mathrm{softplus}(\gamma w_j), \\
  \label{eq:pseudo_posterior}
  \tilde{P}(d_i \mid P,A) &= \frac{\exp(R_i/\tau)}{\sum_{\ell}\exp(R_\ell/\tau)}.
  \end{align}
  Here $\phi_{ij}=\mathrm{softplus}(\psi_{ij})$ is a learnable edge strength with free parameter $\psi_{ij}$, $\gamma>0$ is a learnable absence scale, and $\tau$ is a temperature.
  Eq.~\ref{eq:diag_score} is an additive Gibbs energy over knowledge-base incidences. Positive findings contribute $\log w_j+\log\phi_{ij}$ on the log scale. Confirmed absences contribute a softplus penalty whose magnitude grows with IDF.
  Unknown evidences contribute exactly zero. That is the formal split between missingness and negation.
  Treating unknowns as absent would replace the zero contribution with a negative term for every unasked finding. Incomplete intakes would then collapse toward low-mass rankings, and completion would lose its value.
  The softmax in Eq.~\ref{eq:pseudo_posterior} is a ranking score, not a calibrated clinical posterior. An implicit uniform support prior can still bias toward rare diseases when patterns are ambiguous. That bias is acceptable for Hit@$k$ but not for absolute risk reporting.
  
  Training Objective, Gradients, and Curriculum.
  The optimized parameters are $\psi$, which determines $\phi=\mathrm{softplus}(\psi)$ on knowledge-base edges, and the scalar $\gamma$ that scales soft absence penalties.
  We initialize $\phi_{ij}=\mathrm{softplus}(\psi_{ij})$ to $0.78$ on incidence edges and $\gamma$ to $0.15$ by setting $\psi$ and the unconstrained $\gamma$ pre-image to $\log(\mathrm{expm1}(\cdot))$. Unused slots stay at that initialization because gradients are masked by $M$.
  The frozen-init rows in Table~\ref{tab:static} use exactly this start and never update $\psi$ or $\gamma$.
  IDF weights remain frozen functions of $M$.
  Given a minibatch of cases with gold pathology $d^\star$ and randomly masked views of the full evidence set, we minimize listwise cross-entropy
  \begin{equation}
  \label{eq:loss}
  \mathcal{L}(\phi,\gamma)=-\sum_{c}\log \tilde{P}\big(d^\star_c \mid P_c,A_c\big)+\lambda\|\log\phi\|_2^2.
  \end{equation}
  This loss is the negative log-likelihood of a Plackett-Luce style ranking model restricted to the top label. Gradients push mass onto $d^\star$ under the current ternary view. The $\ell_2$ penalty on $\log\phi$ discourages arbitrarily sharp edges.
  Gradients on $\phi$ are masked to incidence edges with $M_{ij}=1$, so unused slots stay at initialization and cannot absorb spurious correlations outside the knowledge base.
  A curriculum ramps the keep-fraction from $0.1$ to $0.5$ over training and, with probability $0.35$, substitutes the released initial-evidence view, so that $\phi$ cannot spend the whole run on nearly complete charts (Appendix~\ref{app:details}).
  Supervision is ranking only. We never back-propagate through gap selection. The trained COTC module can therefore be reused across mass and information-gain policies.
  The corresponding sparse-curriculum training loop is Algorithm~\ref{alg:train} in Appendix~\ref{app:details}.
  
  \subsection{Probabilistic CoT Completion}
  \label{sec:completion}
  
  Given the current masks $P$ and $A$, the COTC module computes $\tilde{P}$ and Shannon entropy $H=-\sum_i \tilde{P}_i\log\tilde{P}_i$.
  Candidate gaps are unknown evidences that are incident to the top-$k$ diseases under $\tilde{P}$.
  
  Mass-Weighted Priority.
  A simple baseline scores gap $e_j$ by posterior mass on diseases that require it,
  \begin{equation}
  \label{eq:mass}
  \pi_{\mathrm{mass}}(e_j)=\sum_{i: M_{ij}=1}\tilde{P}_i\,\phi_{ij},
  \end{equation}
  restricted to slots with $U_j=1$.
  This prefers questions that support currently likely diseases.
  It can be viewed as a first-order score that rewards edges already carrying mass. It does not explicitly maximize expected entropy reduction and can over-ask about findings that confirm the leader without separating near ties.
  
  Information-Gain Surrogate for One-Step Acquisition.
  Let $p_j$ be the current posterior mass on diseases incident to $e_j$, clipped to the interval from $0.05$ to $0.95$ for stability.
  After setting $e_j$ present or absent in a hypothetical update we obtain entropies $H_{\mathrm{yes}}$ and $H_{\mathrm{no}}$ by recomputing Eqs.~\ref{eq:diag_score} and~\ref{eq:pseudo_posterior} under the two counterfactual masks.
  The surrogate entropy reduction is
  \begin{equation}
  \label{eq:ig}
  \mathrm{IG}(e_j)= H - \big(p_j H_{\mathrm{yes}} + (1-p_j) H_{\mathrm{no}}\big).
  \end{equation}
  Eq.~\ref{eq:ig} is a one-step acquisition heuristic under the ranking scores $\tilde{P}$. It is not Bayesian mutual information because $p_j$ is not a calibrated symptom likelihood. The agent asks the unknown whose two counterfactual updates have the largest weighted entropy reduction under the same energy used for final ranking.
  Using $p_j$ as the Bernoulli weight is an explicit approximation when a generative likelihood for answering yes is unavailable. Clipping avoids degenerate extremes when the shortlist is already peaked. We retain the short label COTC-IG for this surrogate throughout the experiments.
  In practice we evaluate Eq.~\ref{eq:ig} over a shortlist of mass-priority candidates, namely the top-$m$ gaps under Eq.~\ref{eq:mass}.
  A counterfactual rescore of one gap costs $O(|D|)$, so the whole shortlist costs $O(m\,|D|)$ in total rather than a sweep over every unknown evidence.
  Because the question policy is deterministic given $(\phi,\gamma)$, gains from training and gains from information-gain selection can be ablated separately.
  
  \begin{algorithm}[t]
  \caption{Bounded probabilistic CoT completion in COTCAgent}
  \label{alg:cotc_loop}
  \begin{algorithmic}[1]
  \Require KB $M$, COTC parameters $(\phi,\gamma,\tau)$, initial $(P,A)$, threshold $\theta$, max rounds $R_{\max}$, asks per round $a$, shortlist size $m$
  \State Initialize evidence log $\mathcal{L}\gets\emptyset$
  \For{$r = 1$ to $R_{\max}$}
    \State Compute $\{R_i\}$, $\tilde{P}$, $H$ via Eqs.~\ref{eq:diag_score} and~\ref{eq:pseudo_posterior}
    \If{$\max_i \tilde{P}_i \ge \theta$ \textbf{or} no unknown gaps remain}
      \State \textbf{return} ranked diseases, $\mathcal{L}$
    \EndIf
    \State Shortlist top-$m$ gaps by $\pi_{\mathrm{mass}}$; select up to $a$ by $\mathrm{IG}$ via Eq.~\ref{eq:ig}
    \State Ask templated questions; parse answers into present or absent updates
    \State Append updates to $\mathcal{L}$; refresh $(P,A)$
  \EndFor
  \State \textbf{return} ranked diseases, $\mathcal{L}$
  \end{algorithmic}
  \end{algorithm}
  
  Stopping and Oracles.
  Stopping uses the mass threshold $\theta$, an empty gap set, or the round limit $R_{\max}$.
  Questions are templates over evidence names. Answers map to present or absent labels that are written into the evidence log $\mathcal{L}$.
  The threshold $\theta$ trades early stopping against leftover uncertainty. Lower values save questions but risk locking onto a peaked yet incorrect leader. Values near $0.85$ sit at the empirical knee of Hit@1 versus cost on DDXPlus cold-start.
  An oracle that asks only among gold-positive evidences upper-bounds policy quality under perfect evidence selection.
  Algorithm~\ref{alg:cotc_loop} shares one COTC energy for ranking, gap scoring, and re-scoring after each answer. That shared energy is probabilistic CoT completion.
  All hyperparameters $\theta$, $\tau$, $R_{\max}$, $a$, and $m$ are fixed at validation time and held constant across matched-budget comparisons unless a sweep is explicitly reported.
  Because answers enter $\mathcal{L}$ as discrete tokens, every ranked list can be replayed from the knowledge-base energy without calling a language model again. That is what makes the question budget and the ranked list jointly checkable.
  
  \section{Experiments}
  \label{sec:experiments}
  
  \subsection{Experimental Settings}
  \label{sec:settings}
  
  DDXPlus~\cite{fansi2022ddxplus} is the primary benchmark. It provides structured differentials with initial evidence, full evidence sets, and gold pathology.
  We train $\phi$ and $\gamma$ on up to 200k rows with the sparse curriculum in Algorithm~\ref{alg:train}, report static Hit@$k$ on 8k held-out cases, and run consultation on 1k cold-start cases from initial evidence only.
  Unless noted otherwise, consultation uses five rounds and two questions per round with early-stop mass $\theta = 0.85$. Every policy shares this checkpoint, the gold present-or-absent oracle, and the stopping rules.
  MIMIC-IV v3.1~\cite{johnson2023mimic} is the secondary check. Admission labs become discrete tokens and multi-label phenotype groups, the split is patient-disjoint, and the knowledge base is counted on the training split only. Hit@1 asks whether the top phenotype lies in the gold set. This is a completion check under lab sparsity, not a bedside diagnosis benchmark.
  The 8k static initial-evidence Hit@1 of 36.3\% and the 1k no-question Hit@1 of 37.9\% come from different slices. Hyperparameters are in Appendix~\ref{app:details}.
  Baselines.
  Internal policies are no questions, random unknowns, frequency and IDF, mass-weighted gaps, information-gain gaps, and a gold-evidence oracle.
  Under the same budget we reimplement a BASD-style MLP~\cite{fansi2022ddxplus,luo2020asd}, an AARLC-style entropy stopper~\cite{yuan2021aarlc}, a Diaformer-style Transformer~\cite{chen2022diaformer}, fixed-KB information gain~\cite{guan2021bed}, and Qwen2.5-Coder 7B as a local ask-and-rank model~\cite{wang2024agentbench,nori2025sequential}. These are protocol-matched copies, not official checkpoints. COTC and the learned askers share the 200k-row pool, the train/validation split, and 14 epochs. The 7B model is used because it emits parseable evidence IDs (Appendix~\ref{app:ext_baselines}).
  We report Hit@1 and average questions over seeds $0$, $1$, and $2$. Deterministic policies on the 1k slice have seed standard deviation $0$. Random questioning is $56.3\%\pm 0.4$ Hit@1, and the gold-evidence oracle is $98.4\%\pm 0.3$.
  
  \subsection{Comparison Results}
  \label{sec:results}
  
  With the learned COTC parameters, full evidence yields 98.8\% Hit@1, 99.5\% Hit@2, and 99.9\% Hit@3 on 8k DDXPlus cases (Table~\ref{tab:static}). Partial present-only evidence yields 90.0\% Hit@1. Initial evidence alone yields 36.3\% Hit@1.
  Freezing $\phi$ at initialization drops full Hit@1 to 71.9\%, so training the COTC edge matrix is necessary for the ranking layer.
  The gap between full evidence and initial evidence, 98.8\% versus 36.3\%, is the room that completion must recover under cold start.
  
  \begin{table}[t]
  \centering
  \caption{DDXPlus static ranking (\%) under different evidence completeness ($n{=}8000$).}
  \label{tab:static}
  \small
  \renewcommand{\arraystretch}{1.08}
  \begin{tabular}{llcccc}
  \toprule
  \rowcolor{tabheader}
  Model & Evidence & Hit@1 & Hit@2 & Hit@3 & Hit@5 \\
  \midrule
  \rowcolor{tabbest}
  Learned $\phi$ & Full & 98.8 & 99.5 & 99.9 & 100.0 \\
  \rowcolor{tabalt}
  Learned $\phi$ & Partial & 90.0 & 95.6 & 97.4 & 98.8 \\
  Learned $\phi$ & Initial & 36.3 & 50.6 & 59.6 & 72.5 \\
  \rowcolor{tabalt}
  Frozen init $\phi$ & Full & 71.9 & 80.7 & 83.1 & 86.0 \\
  Frozen init $\phi$ & Partial & 63.4 & 72.4 & 76.3 & 80.1 \\
  \rowcolor{tabalt}
  Frozen init $\phi$ & Initial & 25.5 & 35.2 & 40.9 & 52.8 \\
  \bottomrule
  \end{tabular}
  \end{table}
  
\begin{figure}[t]
    \centering
    \includegraphics[width=\linewidth]{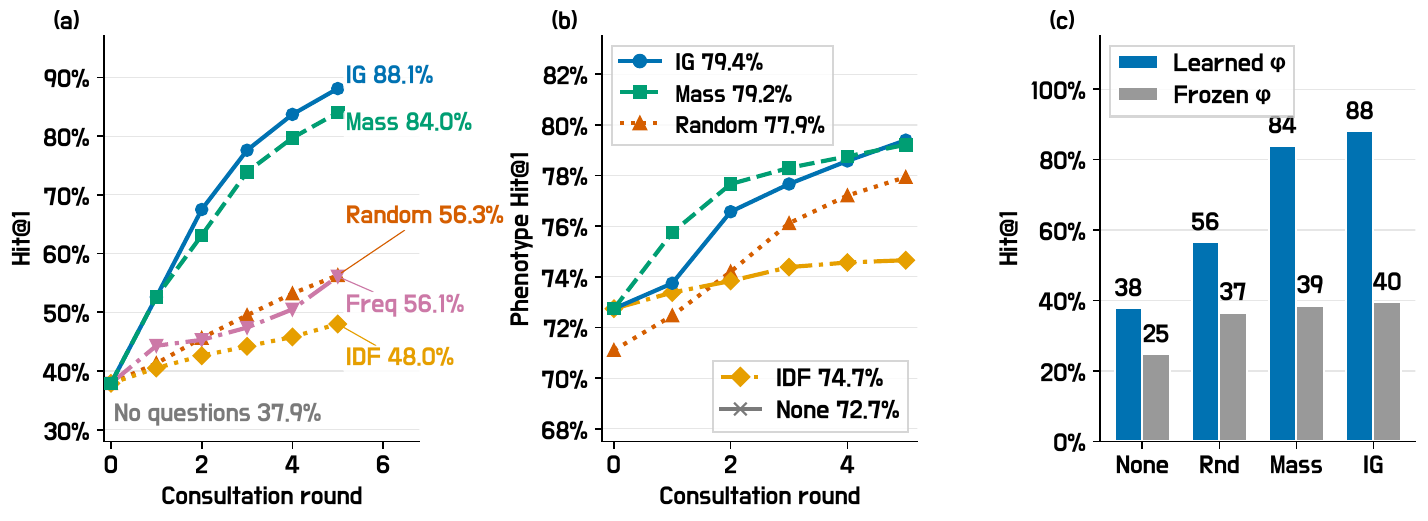}
    \caption{Main consultation results.
    (a)~DDXPlus cold-start Hit@1 by round under a matched budget.
    (b)~MIMIC-IV phenotype Hit@1 by round.
    (c)~Learned versus frozen $\phi$ under matched DDXPlus policies. Bar heights are Hit@1.}
    \label{fig:main_results}
  \end{figure}

  Table~\ref{tab:matched} and Figure~\ref{fig:main_results} summarize cold-start consultation under the matched budget.
  Information gain reaches 88.1\% Hit@1 with 3.61 questions. It beats mass at 84.0\% with 3.86 questions, random at 56.3\% with 5.45 questions, frequency at 56.1\%, and IDF at 48.0\%, while asking fewer questions than the non-adaptive baselines.
  An oracle that samples only gold evidences reaches 98.4\% with 2.18 questions. Remaining room therefore sits in gap selection rather than in the trained COTC energy alone. Under perfect evidence choice the same $R_i$ nearly recovers full-evidence static accuracy.
  
  \begin{table}[t]
  \centering
  \caption{DDXPlus cold-start consultation under a matched budget with 5 rounds and 2 questions per round ($n{=}1000$). Hit@1 (\%) and average questions.}
  \label{tab:matched}
  \small
  \renewcommand{\arraystretch}{1.08}
  \begin{tabular}{lcc}
  \toprule
  \rowcolor{tabheader}
  Policy & Hit@1 & Avg.\ questions \\
  \midrule
  None (no questions) & 37.9 & 0.00 \\
  \rowcolor{tabalt}
  Random & 56.3 & 5.45 \\
  Frequency & 56.1 & 5.49 \\
  \rowcolor{tabalt}
  IDF & 48.0 & 5.78 \\
  COTC-mass & 84.0 & 3.86 \\
  \rowcolor{tabbest}
  COTC-IG & 88.1 & 3.61 \\
  \rowcolor{tabalt}
  Oracle gold evidences & 98.4 & 2.18 \\
  \bottomrule
  \end{tabular}
  \end{table}
  
  Table~\ref{tab:ext_baselines} places COTC against protocol-matched copies of recent acquisition methods on one shared cold-start slice of $n{=}1000$.
  BASD-MLP and AARLC-Ent should be read against their own intake ranking of 26.2\% Hit@1, not against COTC's no-question score of 37.9\%. Asking raises that dual-head classifier only to 31.1\% (BASD) or 28.0\% (AARLC, entropy stop at 1.85 questions). The gain is small, and neither matches the learned COTC energy.
  Diaformer-Lite improves modestly. Fixed-KB BED-IG reaches 60.4\% Hit@1 while asking more than COTC.
  A local LLM that both asks and ranks under the same gold answer oracle is weakest at 14.0\% Hit@1 on $n{=}100$ (seed-0 shuffle of the first 100 cases of that slice).
  On those same 100 cases, no-question COTC reaches 32.0\%, COTC-mass 78.0\% with 4.60 questions, and COTC-IG 87.0\% with 4.22 questions. The LLM gap is therefore not an artifact of comparing $n{=}100$ against $n{=}1000$.
  Table~\ref{tab:ext_baselines} uses one shared slice of 1000 cases so that every external method sees the same patients. On that slice COTC-mass reaches 81.2\% Hit@1 with 4.12 questions and COTC-IG reaches 86.2\% with 3.87 questions, close to 84.0\% and 88.1\% on the main cold-start slice in Table~\ref{tab:matched}. The gap is between two slices of the same pool, not a second consultation protocol.
  
  \begin{table}[t]
  \centering
  \caption{Protocol-matched external baselines on DDXPlus cold-start (5 rounds, 2 questions per round).
  BASD, AARLC, Diaformer, BED, and COTC rows use $n{=}1000$. LLM-Ask uses Qwen2.5-Coder 7B on a seed-0 shuffle of 100 cases from the same test pool.
  On that 100-case subset COTC-IG reaches 87.0\% Hit@1 (4.22 questions) versus LLM-Ask 14.0\% (6.59 questions).}
  \label{tab:ext_baselines}
  \small
  \renewcommand{\arraystretch}{1.08}
  \begin{tabular}{lcc}
  \toprule
  \rowcolor{tabheader}
  Method & Hit@1 (\%) & Avg.\ questions \\
  \midrule
  BASD-MLP & 31.1 & 4.73 \\
  \rowcolor{tabalt}
  AARLC-Ent & 28.0 & 1.85 \\
  Diaformer-Lite & 41.6 & 4.00 \\
  \rowcolor{tabalt}
  BED-IG (fixed KB) & 60.4 & 5.65 \\
  LLM-Ask (Qwen2.5-Coder 7B) & 14.0 & 6.59 \\
  \rowcolor{tabalt}
  COTC-mass & 81.2 & 4.12 \\
  \rowcolor{tabbest}
  COTC-IG (Ours) & 86.2 & 3.87 \\
  \bottomrule
  \end{tabular}
  \end{table}
  
  Figure~\ref{fig:main_results}(a) shows that information gain pulls ahead within the first few rounds and keeps a gap over mass thereafter. Random and frequency baselines plateau far below even though they spend more questions.
  Figure~\ref{fig:main_results}(c) shows that the same policies collapse without learned $\phi$. Completion cannot make up for an untrained COTC module.
  On the 1097 test admissions from MIMIC-IV in Figure~\ref{fig:main_results}(b), no questions yield 72.7\% Hit@1, information gain reaches 79.4\% with 2.41 questions, mass reaches 79.2\% with 2.34 questions, and random reaches 77.9\% with 2.58 questions.
  Mass and information gain are nearly tied here under coarser phenotype labels and denser lab co-occurrence than in DDXPlus. We treat MIMIC as a secondary check under real EHR noise, not as a prospective diagnosis claim.
  
  Training is necessary (Figure~\ref{fig:main_results}(c)); intake fraction, stopping, and shortlist results are reported with question costs in Appendix Figures~\ref{fig:budget}, \ref{fig:train_intake}, and~\ref{fig:dropout}, Tables~\ref{tab:budget}--\ref{tab:frozen}, and Appendix~\ref{app:llm}.
  
  \subsection{Discussion}
  \label{sec:discussion}
  
  The results support three scoped conclusions. First, ternary unknown states preserve the value of asking by distinguishing unobserved findings from confirmed absences. Collapsing these states would prematurely penalize plausible diseases, distort the initial ranking, and reduce the usefulness of follow-up questions. Second, learned edge strengths and gap selection are complementary: the former determines how observed evidence changes disease scores, whereas the latter determines which unresolved finding is most useful to acquire next. Neither component alone is sufficient—question selection cannot compensate for an inaccurate disease–evidence structure, while a well-trained scoring model cannot improve efficiently if it queries uninformative findings. Third, interactive methods must be compared under matched answer oracles, consultation rounds, question limits, and training-data exposure, because accuracy alone may conceal substantially different evidence costs or supervision advantages.
  The matched DDXPlus numbers make that split concrete. Information gain reaches 88.1\% Hit@1 in 3.61 questions, while the gold-evidence oracle reaches 98.4\% in 2.18 questions, so most of the remaining error is which unknown to ask. Frequency and IDF spend more questions, 5.49 and 5.78, yet stay near or below random at 56.1\% and 48.0\%. Prevalence and rarity therefore do not replace the learned scores. With frozen edges, the same information-gain rule reaches only 39.6\% Hit@1 and uses 6.80 questions, close to the learned no-question score of 37.9\%. AARLC stops earlier, at 1.85 questions and 28.0\% Hit@1, so a smaller question count is not by itself a better consultation.
  
  MIMIC-IV remains a secondary phenotype-ranking robustness check rather than a prospective diagnostic evaluation. TSA actively converts its irregular chemistry series into abnormality and trend predicates used by COTC. However, the coarser phenotype labels, denser laboratory co-occurrence, and retrospective observation process reduce the separation between mass- and information-gain policies. Their near tie should therefore be interpreted as evidence that policy advantages depend on label granularity and evidence structure, not as evidence that TSA is inactive or unnecessary. Appendix~\ref{app:discussion} provides the complete interpretation.
  
  The method nevertheless assumes that answers can be mapped reliably to present or absent states, produces ranking scores rather than calibrated posterior probabilities, and remains an in-knowledge-base model. Performance on held-out diseases falls near random (Appendix~\ref{app:holdout}), indicating that information-gain questioning cannot recover disease–evidence relations that were never learned. The automatic answer oracle also removes linguistic ambiguity, patient refusal, measurement error, and real consultation costs. Consequently, the reported gains isolate the structured completion mechanism under controlled conditions; they do not establish open-world generalization, calibrated clinical risk, patient safety, or clinical utility.
  
  \section{Conclusion}
  
  COTCAgent casts preventive consultation as bounded probabilistic completion of a ternary evidence log under a public question budget.
  Separating TSA, a learnable knowledge-base energy, and deterministic gap selection keeps rankings tied to evidence that can be replayed rather than to free-form chain of thought.
  Training and information-gain completion are complementary. Structured completion can stay competitive without an LLM ranker under matched budgets.
  Calibrated scoring and prospective evaluation are left to future work.

\bibliographystyle{iclr2027_conference}
\hypersetup{urlcolor=black,linkcolor=black}
\bibliography{main}
\hypersetup{urlcolor=blue,linkcolor=blue}

\clearpage
\appendix
\setcounter{table}{0}
\setcounter{figure}{0}
\setcounter{algorithm}{0}
\renewcommand{\thetable}{A\arabic{table}}
\renewcommand{\thefigure}{A\arabic{figure}}
\renewcommand{\thealgorithm}{A\arabic{algorithm}}
\providecommand{\theHtable}{\thetable}
\providecommand{\theHfigure}{\thefigure}
\renewcommand{\theHtable}{A.\arabic{table}}
\renewcommand{\theHfigure}{A.\arabic{figure}}

\begin{center}
{\Large\bfseries Supplementary Material}\\[0.35em]
{\large Appendix}
\end{center}
\vspace{0.8em}

\section{Additional experimental details}
\label{app:details}

This appendix expands protocols that the main text summarizes only briefly, so that numbers, splits, and stopping rules can be reproduced without ambiguity.

\noindent\textbf{Training.}
The reported DDXPlus checkpoint is \texttt{coldstart\_curriculum}.
It is trained on 200k training rows for 14 epochs with learning rate $0.015$ under a cosine schedule, batch size 512, and weight decay $10^{-4}$.
Sparse masking uses absent fraction $0.25$ and initial-mix probability $0.35$.
When the curriculum flag is on, the keep-fraction ramps from $0.1$ to $0.5$ across epochs rather than staying at the nominal $0.25$; with probability $0.35$ a minibatch case is replaced by its released initial-evidence view, otherwise it is a randomly thinned view of the full evidence set.
Optimization uses listwise cross-entropy on the gold pathology as in Eq.~\ref{eq:loss} of the main text.
We select the checkpoint by a composite of cold-start information-gain Hit@1, question cost, and static partial or initial scores among parallel training configurations, as shown in Figure~\ref{fig:config}.

\begin{algorithm}[ht]
\caption{Sparse-curriculum training of the COTC parameters $(\phi,\gamma)$}
\label{alg:train}
\begin{algorithmic}[1]
\Require Cases $\{(E^{\mathrm{full}}_c,d^\star_c)\}$, KB $M$, epochs $E$, keep/absent schedules, initial-mix probability $p_{\mathrm{mix}}$
\State Initialize $\psi_{ij}$ so that $\phi_{ij}=0.78$ on edges with $M_{ij}=1$; $\gamma\leftarrow 0.15$
\For{$e=1$ to $E$}
  \State Sample minibatch; draw keep fraction $k_e$ and absent fraction $a_e$
  \For{each case $c$}
    \If{$\mathrm{Unif}(0,1)<p_{\mathrm{mix}}$}
      \State Use the released initial-evidence view of $c$
    \Else
      \State Sample present mask $P_c$ from $E^{\mathrm{full}}_c$ with keep $k_e$; set $A_c$ on a fraction $a_e$ of remaining known negatives
    \EndIf
    \State Compute $R_i$ and $\tilde{P}$ via Eqs.~\ref{eq:diag_score} and~\ref{eq:pseudo_posterior}; accumulate Eq.~\ref{eq:loss}
  \EndFor
  \State Update $\psi,\gamma$ by AdamW; set $\phi=\mathrm{softplus}(\psi)$
\EndFor
\State \textbf{return} $(\phi,\gamma)$ selected by validation composite below
\end{algorithmic}
\end{algorithm}

\noindent\textbf{Consultation protocol.}
Unless stated otherwise, cold-start cases begin with only the released initial evidence and treat every other finding as unknown rather than absent.
Each round asks up to two unknown evidences.
Answers follow the gold evidence set: a finding is labeled present if it appears for the case and absent otherwise.
Early stopping uses mass threshold $\theta{=}0.85$ when enabled, and the sweeps below vary that threshold while holding the ask and round budget fixed.
All policies share the same learned checkpoint and the same answer oracle, so differences in Hit@1 and average questions are attributable to gap selection rather than to inconsistent answering.

\noindent\textbf{Why static and consultation Hit@1 differ at intake.}
Table~\ref{tab:static} in the main text reports initial-evidence Hit@1 of 36.3\% on an 8k static slice, whereas Table~\ref{tab:matched} reports no-question Hit@1 of 37.9\% on a 1k cold-start consultation subset.
These are not contradictory: they evaluate different case draws under the same trained energy.
Throughout the paper we therefore treat static ranking as a completeness probe and consultation as the matched-budget interactive evaluation, and we avoid equating the two intake numbers.

\noindent\textbf{MIMIC construction.}
We sample 6000 admissions from MIMIC-IV v3.1, keep a fixed panel of 18 common lab item IDs (creatinine, BUN, glucose, electrolytes, CBC, anion gap, albumin, bilirubin, lactate, troponin T, INR, PTT), bin values into discrete abnormal or trend tokens, and define 8 multi-label phenotype groups from diagnosis-code chapters (circulatory, endocrine, genitourinary, infectious, respiratory, digestive, neoplasm, injury).
After requiring at least three lab points, 4552 admissions remain. Requiring at least one of the eight theme phenotypes then leaves 4384 admissions.
Train and test splits are patient-disjoint on \texttt{subject\_id} (3287 / 1097 in the reported run) so that no patient contributes both ranking supervision and test labels.
The knowledge base is phenotype and evidence co-occurrence estimated on the training patients only.
The reported metric asks whether the top-1 phenotype lies in the gold phenotype set for that admission, which is a completion check under sparse labs rather than a claim of primary diagnosis accuracy.
The consultation numbers in the main text (no questions 72.7\%, information gain 79.4\% at 2.41 questions) use this patient-disjoint, train-only knowledge-base protocol.

\section{Budget and stopping sweeps}
\label{app:budget}

\begin{figure}[ht]
  \centering
  \resizebox{0.8\linewidth}{!}{%
  \begin{minipage}{\linewidth}
  \centering
  \begin{minipage}[t]{0.48\linewidth}
    \centering
    \includegraphics[width=\linewidth]{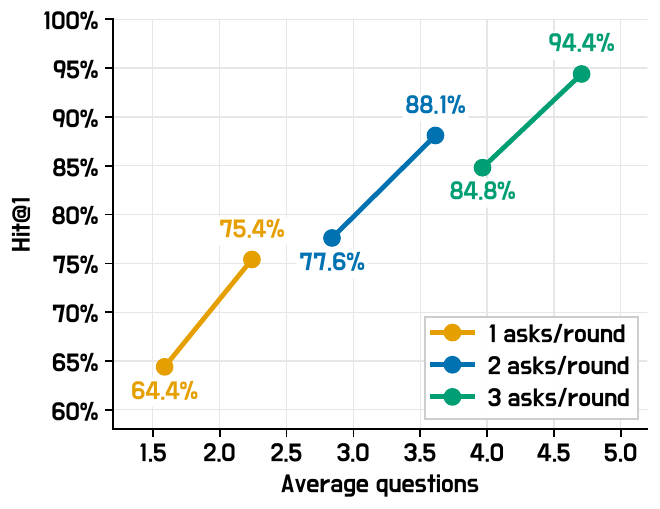}\\
    {\footnotesize (a) Asks/rounds budget Pareto}
  \end{minipage}\hfill
  \begin{minipage}[t]{0.48\linewidth}
    \centering
    \includegraphics[width=\linewidth]{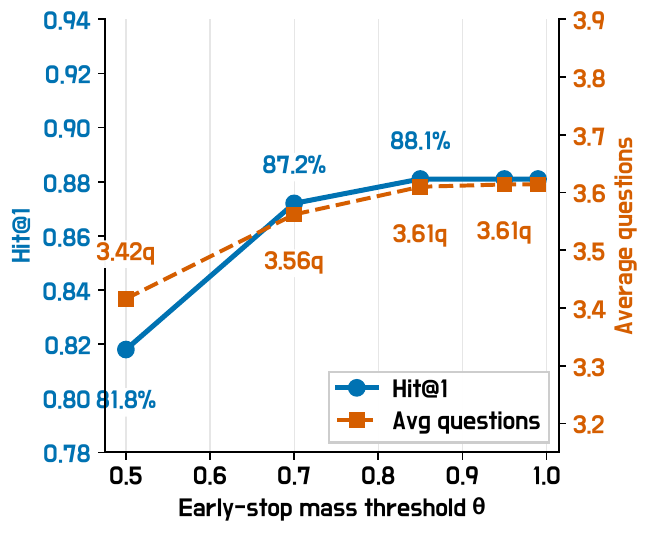}\\
    {\footnotesize (b) Early-stop threshold $\theta$}
  \end{minipage}
  \vspace{0.15em}
  \includegraphics[width=0.48\linewidth]{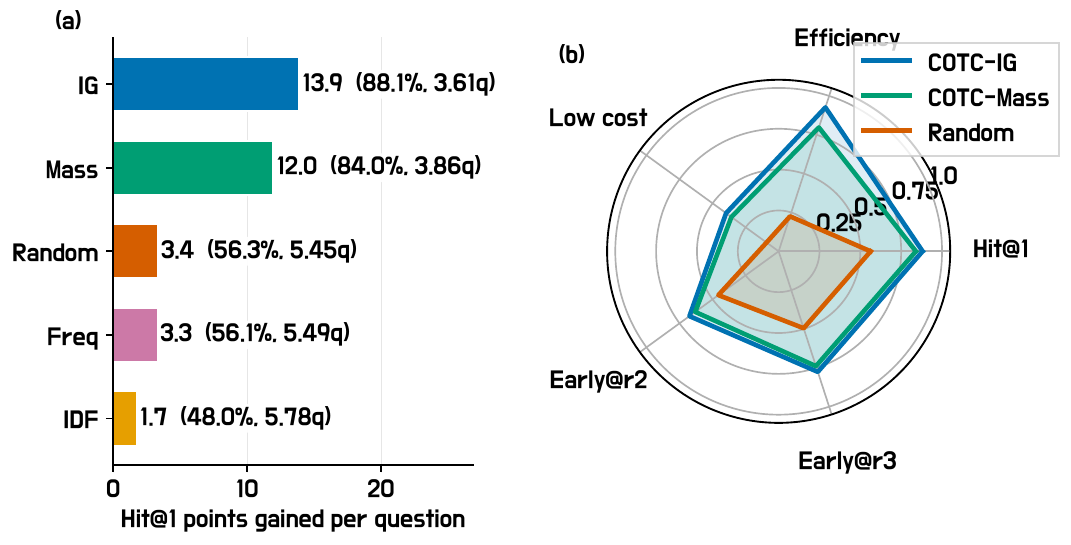}\\
  {\footnotesize (c) Policy efficiency}
  \end{minipage}%
  }
  \caption{Budget, stopping, and efficiency on DDXPlus cold-start.
  (a)~Asks/rounds Pareto (Table~\ref{tab:budget}).
  (b)~Early-stop $\theta$ (Table~\ref{tab:theta}).
  (c)~Hit@1 points per question and multi-metric radar; information gain leads.}
  \label{fig:budget}
  \label{fig:efficiency}
\end{figure}

\begin{figure}[ht]
  \centering
  \includegraphics[width=\linewidth]{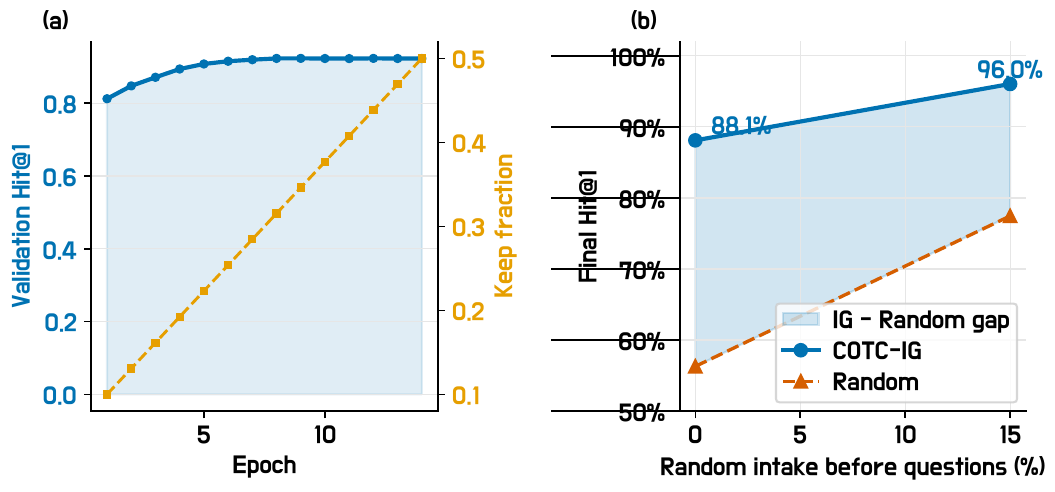}
  \caption{Training dynamics and intake ablation.
  (a)~Cold-start curriculum: validation Hit@1 and keep fraction over epochs.
  (b)~Final Hit@1 versus random intake fraction for COTC-IG and random questioning.}
  \label{fig:train_intake}
\end{figure}

\begin{figure}[ht]
  \centering
  \includegraphics[width=0.72\linewidth]{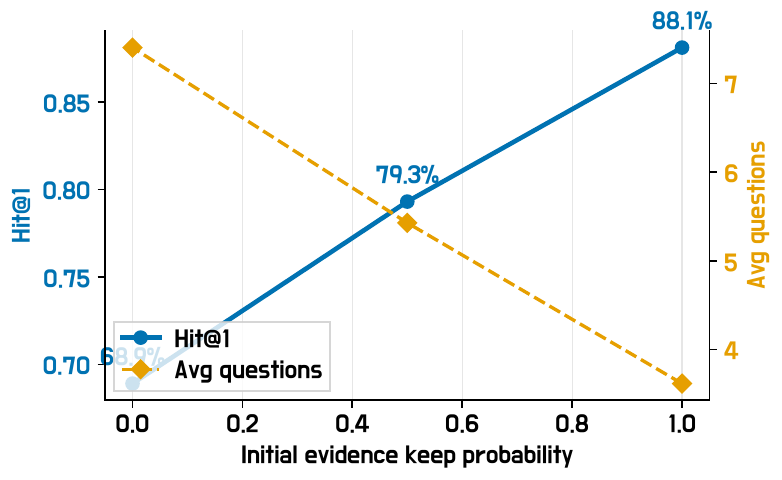}
  \caption{Initial-evidence keep probability stress test under COTC-IG, with Hit@1 on the left axis and average questions on the right axis.}
  \label{fig:dropout}
\end{figure}

The main text reports one operating point with two asks and five rounds at $\theta{=}0.85$.
Figure~\ref{fig:budget} collects the asks/rounds Pareto, the $\theta$ sweep, and the efficiency view; numeric cells remain in Tables~\ref{tab:budget} and~\ref{tab:theta}.
Table~\ref{tab:dropout} randomly drops initial evidences to stress sparse intakes: keep probability $1.0$ recovers the main-text cold start, keep probability $0.5$ removes half of the released intake at random, and keep probability $0.0$ starts from a fully empty evidence log.
Round-wise and learning results appear in main-text Figure~\ref{fig:main_results}.
Figure~\ref{fig:dropout} plots the keep-probability stress test, and Figure~\ref{fig:train_intake} shows curriculum training together with intake ablations.

Reading Table~\ref{tab:budget} row by row clarifies the Pareto claim.
Moving from one ask and three rounds to two asks and five rounds raises Hit@1 from 64.4\% to 88.1\% while average questions rise from 1.59 to 3.61.
Adding a third ask under five rounds further raises Hit@1 to 94.4\% at 4.71 questions, which is useful when a higher budget is available but is no longer the matched setting used for policy comparisons in Table~\ref{tab:matched}.
Table~\ref{tab:theta} shows that lowering $\theta$ to $0.50$ stops earlier and loses accuracy, whereas raising $\theta$ above $0.85$ no longer changes Hit@1 or question cost on this checkpoint, which is why $0.85$ is used as the default operating point.

\begin{table}[ht]
\centering
\caption{COTC-IG budget sweep on DDXPlus cold-start.}
\label{tab:budget}
\begin{tabular}{ccrr}
\toprule
Asks/round & Rounds & Hit@1 (\%) & Avg.\ questions \\
\midrule
1 & 3 & 64.4 & 1.59 \\
1 & 5 & 75.4 & 2.24 \\
2 & 3 & 77.6 & 2.84 \\
2 & 5 & 88.1 & 3.61 \\
3 & 3 & 84.8 & 3.97 \\
3 & 5 & 94.4 & 4.71 \\
\bottomrule
\end{tabular}
\end{table}

\begin{table}[ht]
\centering
\caption{Early-stop mass threshold $\theta$ with 2 asks and 5 rounds.}
\label{tab:theta}
\begin{tabular}{crr}
\toprule
$\theta$ & Hit@1 (\%) & Avg.\ questions \\
\midrule
0.50 & 81.8 & 3.42 \\
0.70 & 87.2 & 3.56 \\
0.85 & 88.1 & 3.61 \\
0.95 & 88.1 & 3.61 \\
0.99 & 88.1 & 3.61 \\
\bottomrule
\end{tabular}
\end{table}

\begin{table}[ht]
\centering
\caption{Initial-evidence keep probability under COTC-IG.}
\label{tab:dropout}
\begin{tabular}{crr}
\toprule
Keep prob. & Hit@1 (\%) & Avg.\ questions \\
\midrule
1.0 & 88.1 & 3.61 \\
0.5 & 79.3 & 5.43 \\
0.0 & 68.9 & 7.41 \\
\bottomrule
\end{tabular}
\end{table}

\section{Policy efficiency and early-stop threshold}
\label{app:efficiency}
\label{app:theta_fig}
\label{app:theta}

Figure~\ref{fig:budget}(b) sweeps the early-stop mass threshold $\theta$ under two asks and five rounds; Table~\ref{tab:theta} lists the numeric cells.
Hit@1 rises from 81.8\% at $\theta{=}0.5$ to 88.1\% at $\theta{=}0.85$ with little extra question cost and then plateaus, which is why the main text uses $\theta{=}0.85$ as the default operating point.
Figure~\ref{fig:budget}(c) reports Hit@1 points gained per question and a multi-metric radar on the same matched budget; information gain leads on both views, consistent with selecting gaps that sharpen $\tilde{P}$ under Eq.~\ref{eq:ig}.

\section{Learned versus frozen consultation}
\label{app:frozen}

\begin{figure}[ht]
  \centering
  \includegraphics[width=\linewidth]{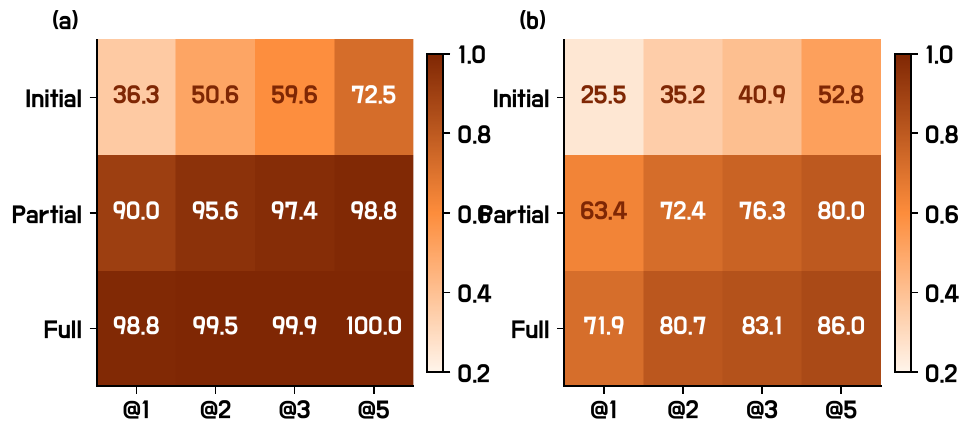}
  \caption{Static Hit@$k$ heatmaps for learned $\phi$ in panel (a) and frozen initialization in panel (b) across evidence completeness.}
  \label{fig:static_hm}
\end{figure}

The main text argues that completion and learning interact.
Table~\ref{tab:frozen} makes that claim quantitative by comparing the learned checkpoint to a frozen-$\phi$ initialization under the same policies, with the bar view in main-text Figure~\ref{fig:main_results}(c).
Without training, information gain only reaches 39.6\% Hit@1 and spends more questions than the learned run, which shows that gap selection cannot recover from an untrained energy.
The learned rows are consistent with the main matched-budget table up to ordinary single-run rounding: the enrichment dump used for the frozen ablation reports Random at 56.5\% Hit@1, whereas the multiseed mean reported in the main text is 56.3\%.
Mass and information-gain rows match exactly across the two tables.
Figure~\ref{fig:static_hm} expands static Hit@$k$ for both models as heatmaps so that the ranking-layer gap is visible before any questioning begins.

\begin{table}[ht]
\centering
\caption{Cold-start consultation: learned $\phi$ versus frozen initialization.}
\label{tab:frozen}
\begin{tabular}{lccrr}
\toprule
Model & Policy & Hit@1 (\%) & Avg.\ q \\
\midrule
Learned & None (no questions) & 37.9 & 0.00 \\
Learned & Random & 56.5 & 5.44 \\
Learned & COTC-mass & 84.0 & 3.86 \\
Learned & COTC-IG & 88.1 & 3.61 \\
Frozen & None (no questions) & 24.7 & 0.00 \\
Frozen & Random & 36.6 & 6.98 \\
Frozen & COTC-mass & 38.6 & 6.88 \\
Frozen & COTC-IG & 39.6 & 6.80 \\
\bottomrule
\end{tabular}
\end{table}

\section{LLM matched-budget details}
\label{app:llm}

\begin{figure}[ht]
  \centering
  \includegraphics[width=\linewidth]{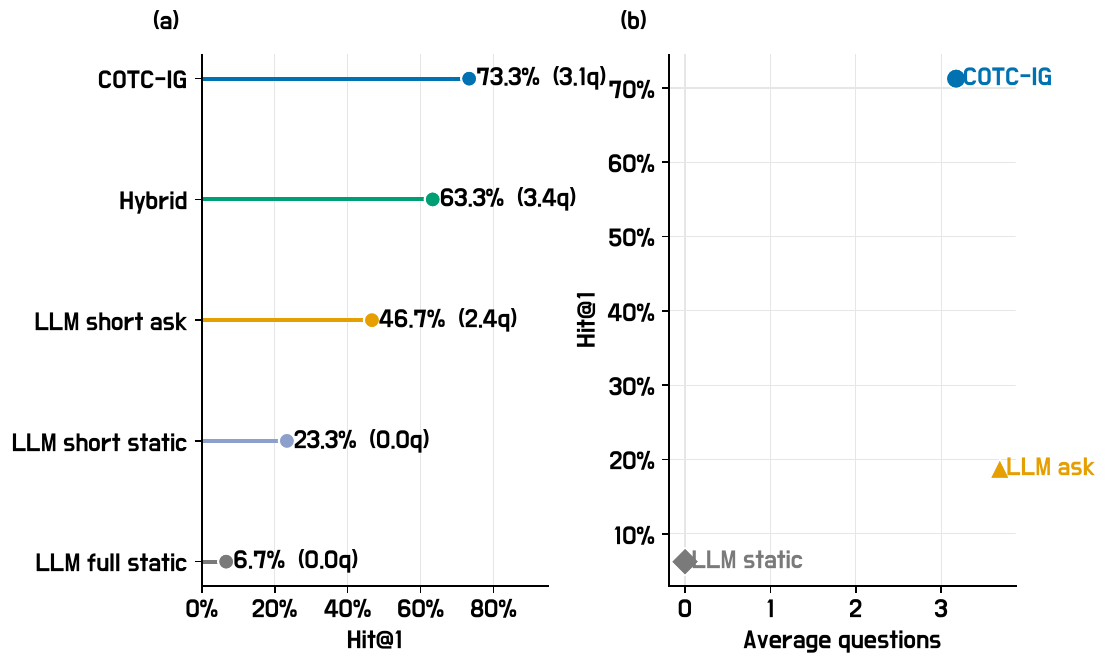}
  \caption{LLM matched-budget comparison.
  (a)~Lollipop Hit@1 on the refined shortlist protocol.
  (b)~Accuracy versus question cost for matched-budget LLM arms.}
  \label{fig:llm}
\end{figure}

The main text reports that structured completion remains competitive without an LLM reader.
This section specifies the protocol so that the comparison is not mistaken for an open-ended dialogue benchmark.
Figure~\ref{fig:llm} summarizes the refined shortlist protocol with $n{=}60$, at most three rounds, two asks per round, and shortlist size 15, together with a matched-budget accuracy and cost scatter.
The language model is Qwen2.5-Coder 7B served locally through Ollama, chosen because the protocol requires parseable evidence IDs rather than fluent counseling text.
All arms share the gold yes-or-no answer oracle used by COTC, so the language model never invents present or absent labels that COTC cannot also receive.
Ranking and question-choice prompts are fixed templates.
The ranker prompt asks for a single numbered disease from a candidate list given present and absent findings and forbids explanation.
The asker prompt lists at most 40 unknown evidences with IDs and English questions and requests $a$ item numbers or \texttt{E\_} identifiers.
The templates used at evaluation time are shown below, with line breaks inserted so that they fit the appendix column.

{\footnotesize
\begin{verbatim}
You are a careful clinical differential-diagnosis assistant.
Given findings, pick the single most likely disease
from the numbered list.
Output format (strict): a single line like
`3. DiseaseName`. Do not explain.

Present findings:
- ...
Absent findings:
- ...
Candidates:
1. ...
Answer:
\end{verbatim}

\begin{verbatim}
Choose the most informative follow-up questions
to refine the differential.
Current findings: ...
Top hypotheses: ...
Pick <a> item numbers from the list below.
Reply with numbers or E_xxx ids only.

1. E_...: <English question>
Picks:
\end{verbatim}
}
COTC-IG ranks with the structured energy only and never queries the language model for ranking.
LLM-full-static ranks among all diseases from the intake text without questions.
LLM-shortlist-static ranks the COTC top-15 without questions, which is a fairer reader baseline because the candidate set is already narrowed.
LLM-shortlist-ask lets the language model choose questions and then rank the shortlist.
Hybrid uses COTC-IG questions and a language-model shortlist ranker, which isolates whether the language model helps mainly at question selection or at final ranking.
Hybrid does \emph{not} use the COTC mass threshold $\theta$: each round the language model re-ranks the current COTC top-$k$, and the loop stops when that pick matches the gold pathology or the round limit fires.
That is why Hybrid asks 3.40 questions on average against 3.13 for COTC-IG on the same shortlist protocol: a wrong LLM pick continues asking even when the structured mass would already have stopped.

Table~\ref{tab:llm} lists the corresponding numbers.
COTC-IG reaches 73.3\% Hit@1 with 3.13 questions.
Hybrid reaches 63.3\% with 3.40 questions, which suggests that replacing the structured ranker with the language-model ranker hurts more than it helps under this protocol.
LLM-shortlist-ask reaches only 46.7\% despite asking fewer questions on average, and full-list static ranking collapses to 6.7\%.
These results support the architectural claim that, when answers are shared, the bottleneck is gap selection and energy scoring rather than free-form generation.
They do not claim that larger proprietary models would fail under a different protocol.

\begin{table}[ht]
\centering
\caption{LLM comparison numbers corresponding to Figure~\ref{fig:llm}(a).}
\label{tab:llm}
\begin{tabular}{lcc}
\toprule
Policy & Hit@1 (\%) & Avg.\ questions \\
\midrule
COTC-IG & 73.3 & 3.13 \\
Hybrid COTC ask, LLM rank & 63.3 & 3.40 \\
LLM shortlist ask & 46.7 & 2.37 \\
LLM shortlist static & 23.3 & 0.00 \\
LLM full static & 6.7 & 0.00 \\
\bottomrule
\end{tabular}
\end{table}

\section{Disease hold-out}
\label{app:holdout}

\begin{figure}[ht]
  \centering
  \includegraphics[width=\linewidth]{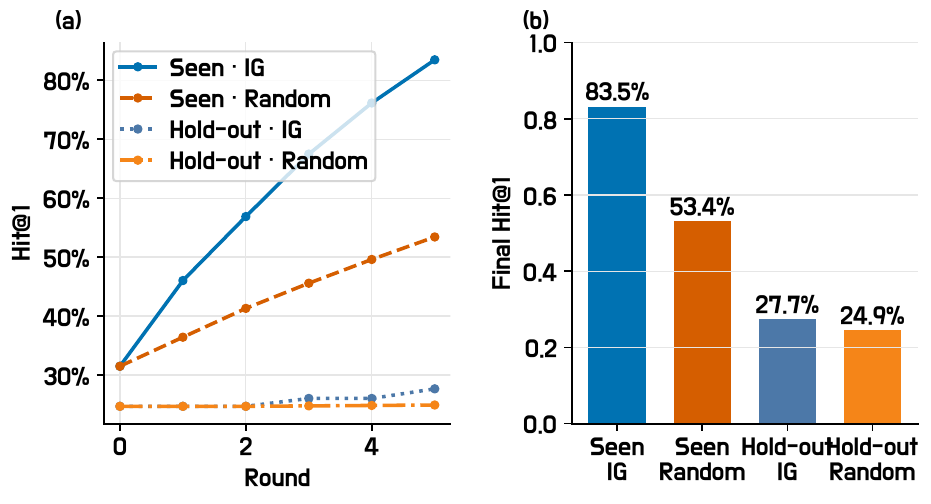}
  \caption{Disease hold-out.
  (a)~Round-wise Hit@1 for seen versus hold-out subsets under information gain and random questioning.
  (b)~Final Hit@1 summary.}
  \label{fig:holdout}
\end{figure}

Figure~\ref{fig:holdout} reports the disease hold-out stress test.
Pathologies withheld from $\phi$ training remain near random under information-gain completion, while seen diseases retain strong gains.
This pattern is expected for an in-knowledge-base completion model: edges that never receive ranking supervision cannot be recovered by questioning alone.
We therefore treat hold-out as a negative control that bounds the claim, not as a failure of the matched-budget result on seen diseases.
The main text does not sell open-world diagnosis, and this appendix is the place where that non-claim is made quantitative.

\section{Training configuration selection}
\label{app:config}

\begin{figure}[ht]
  \centering
  \includegraphics[width=\linewidth]{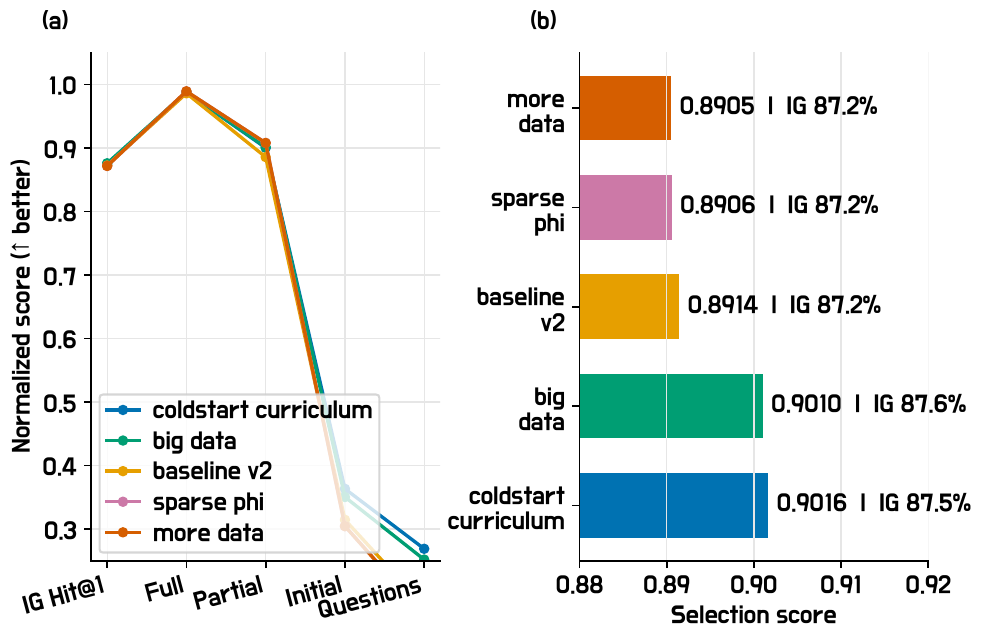}
  \caption{Configuration selection.
  (a)~Parallel-coordinate style profiles with questions inverted so that higher is better.
  (b)~Selection-score ranking with information-gain Hit@1 annotations.}
  \label{fig:config}
\end{figure}

Figure~\ref{fig:config} compares parallel training configurations with normalized metric profiles and selection scores.
\texttt{coldstart\_curriculum} ranks first on the composite used for the main checkpoint.
The unnormalized selection score is
\begin{equation}
\label{eq:select}
s = \mathrm{Hit@1}_{\mathrm{IG}} - 0.01\cdot \overline{q}_{\mathrm{IG}} + 0.05\cdot \mathrm{Hit@1}_{\mathrm{partial}} + 0.05\cdot \mathrm{Hit@1}_{\mathrm{initial}},
\end{equation}
which balances cold-start information-gain Hit@1 against question cost and static partial or initial accuracy.
Other configurations may look stronger on a single axis, for example static full evidence alone, yet lose on cold-start consultation, which is the primary claim setting of this paper.
The figure is therefore a selection audit rather than an exhaustive hyperparameter search paper.

\section{PhysioNet TSA grounding}
\label{app:physio}

\begin{figure}[ht]
  \centering
  \includegraphics[width=\linewidth]{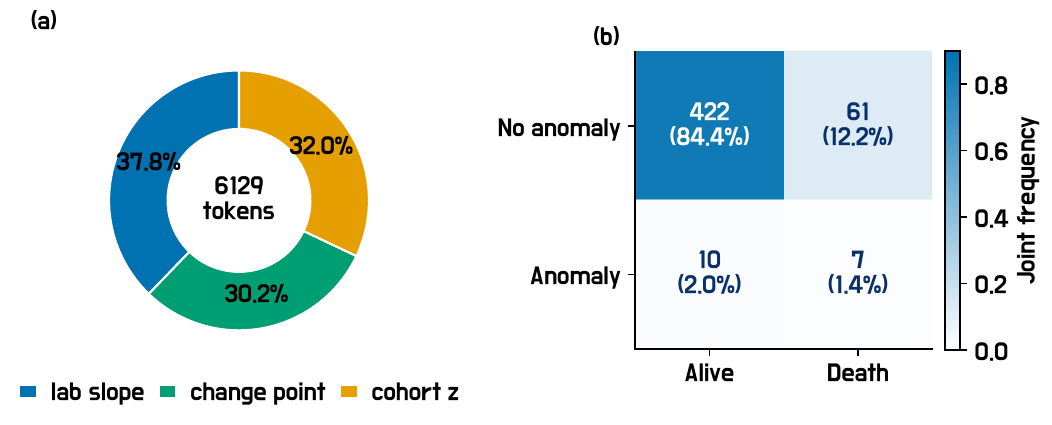}
  \caption{PhysioNet 2012 TSA outputs.
  (a)~Donut of estimand token counts.
  (b)~Joint frequency of anomaly flags and death.}
  \label{fig:physio}
\end{figure}

DDXPlus does not exercise TSA because symptom evidence is already discrete.
To show that the executable-statistics layer is not vacuous when longitudinal inputs exist, Figure~\ref{fig:physio} summarizes TSA estimand composition on PhysioNet 2012 ICU series and a contingency heatmap of cohort-$z$ anomaly flags versus in-hospital mortality.
Panel (a) shows that the adapter emits a mixture of slope, change-point, and cohort-$z$ tokens rather than a single canned phrase; residual RMSE is attached as a \texttt{STABLE}/\texttt{UNSTABLE} quality flag on the slope token, and cohort-$z$ values with $|z|>2.5$ are the anomaly flags in panel (b).
Panel (b) checks that anomaly flags co-occur with mortality in a structured way, which is a grounding sanity check rather than a predictive claim about PhysioNet itself.
This appendix therefore supports the main-text statement that TSA is idle on DDXPlus yet ready when raw series are available, as on MIMIC lab binning.

\section{Extended discussion of secondary settings}
\label{app:discussion}

Main-text Table~\ref{tab:matched} and Figure~\ref{fig:main_results} already give the matched-budget numbers. This section records how those numbers should be read on MIMIC-IV, where TSA is not idle.
DDXPlus findings are already discrete, so TSA does not run. MIMIC chemistry series are irregular, so TSA maps them to abnormality and trend predicates before COTC scores the admission.
The near tie between mass and information gain on MIMIC is therefore not a claim that TSA is unused. Phenotype groups are coarser than DDXPlus pathologies, laboratory tokens co-occur more densely, and the chart is observed retrospectively rather than collected by asking. Under that structure, many remaining gaps change the ranking in similar ways, so the two policies land close together in Hit@1 and question cost.
We still treat MIMIC as a secondary robustness check: Hit@1 asks whether the top phenotype lies in the gold set, not whether a primary bedside diagnosis is recovered. Appendix~\ref{app:nonclaims} lists claims we do not make.

\section{What we do not claim}
\label{app:nonclaims}

We do not claim open-world diagnosis for held-out pathologies; hold-out Hit@1 stays near random because $\phi$ is never trained on those diseases, as shown in Figure~\ref{fig:holdout}.
We do not claim that softmax scores are calibrated clinical probabilities.
We do not use HealthBench or synthetic vignette leaderboards as primary evidence for this paper.
MIMIC phenotype Hit@1 is a secondary completion check, not a substitute for prospective clinical evaluation.
We also do not claim that the local 7B LLM comparison exhausts the space of language-model readers; it only shows that, under a shared answer oracle and matched ask budget, structured completion does not require an LLM ranker to remain competitive.

\section{External baseline reimplementations}
\label{app:ext_baselines}

Table~\ref{tab:ext_baselines} uses protocol-matched reimplementations rather than official author checkpoints, because public interactive DDXPlus releases rarely expose ternary unknown states, matched ask budgets, and gold yes/no answer oracles in one package.
\textbf{BASD-MLP} trains a two-layer MLP with a disease head and a next-evidence head on masked ternary views, then asks the top unknown scores and stops by mass threshold.
\textbf{AARLC-Ent} reuses that backbone but stops when ranking entropy falls below a validation threshold, capturing the entropy-alignment idea of AARLC without full RL rollouts.
\textbf{Diaformer-Lite} replaces the MLP with a two-layer Transformer encoder over present/absent/unknown tokens and the same dual heads.
\textbf{BED-IG} runs information gain on a frozen knowledge-base energy with $\phi{\approx}1$, isolating experimental design without learned edges.
\textbf{LLM-Ask} uses local Qwen2.5-Coder 7B to choose questions and rank diseases under the same gold answers.
All neural baselines are trained on 80k DDXPlus training rows for five epochs and evaluated on 1k cold-start cases with five rounds and two asks; this is a lighter reimplementation budget than the 200k-row, 14-epoch COTC energy, which has far fewer parameters.
LLM-Ask uses $n{=}100$ because each case requires multiple Ollama calls.
On the identical 100-case shuffle, COTC-IG reaches 87.0\% Hit@1 with 4.22 questions (COTC-mass 78.0\% / 4.60; no questions 32.0\%), which is the fair subset comparison against LLM-Ask at 14.0\% / 6.59.
Raw JSON is stored at \texttt{results/baselines/matched\_budget\_baselines.json}.

\end{document}